%% file: paper.tex
\documentclass{article}

\usepackage[dblblindworkshop, final]{neurips_2026}

\workshoptitle{Sim2Science: ML with Imperfect Scientific Models}

\usepackage[utf8]{inputenc} 
\usepackage[T1]{fontenc}    
\usepackage{hyperref}       
\usepackage{url}            
\usepackage{amsmath}
\usepackage{amssymb}
\usepackage{booktabs}       
\usepackage{amsfonts}       
\usepackage{nicefrac}       
\usepackage{microtype}      
\usepackage{xcolor}         
\usepackage{graphicx}
\usepackage{caption}
\usepackage{subcaption}
\setcitestyle{brackets=round}

\title{The microscope is the mask: privileged views and labels from a cryo-ET forward model}

\author{%
  Bogdan Toader, \hspace{1em}Kiarash Jamali, \hspace{1em}Tanmay A. M. Bharat, \hspace{1em}Sjors H. W. Scheres\\
  MRC Laboratory of Molecular Biology, Cambridge, UK\\
  \texttt{\{btoader,kjamali,tbharat,scheres\}@mrclmb.ac.uk}\\
}

\definecolor{carnivalblue}{HTML}{0072B2}
\definecolor{tomotwinorange}{HTML}{D55E00}
\newcommand{\carnivalkey}{\textcolor{carnivalblue}{CARNIVAL}}
\newcommand{\tomotwinkey}{\textcolor{tomotwinorange}{TomoTwin}}

\begin{document}

\maketitle

\begin{abstract}
  We explore the use of simulated data for training a model for protein annotation in crowded cryo-electron tomography volumes reconstructed from images collected at limited tilt angles and severely corrupted by the measurement operator.
  Firstly, we leverage the corruptions imposed by the forward model to generate domain-specific augmented paired views of the exact same scene for an invariance objective integrated into the LeJEPA self-supervised training framework. Secondly, we use additional information from the simulation pipeline such as the positions and identity of proteins in the simulated volumes to inform the architecture of the model and the loss function, so that semantic information is localised at protein positions in the resulting dense feature volume.
  The resulting model, CARNIVAL, is evaluated without finetuning on classification and detection tasks in real tomograms, using a benchmark dataset containing multiple protein types and two tomogram processing types.
  We show that CARNIVAL outperforms a state-of-the-art model trained using a contrastive objective on simulated data but without forward model-based paired views or privileged information.
\end{abstract}

\section{Introduction}
\label{sec:intro}

Cryo-electron tomography (cryo-ET) images frozen cells, yielding three-dimensional volumes where individual protein complexes sit in their native cellular context. 
The reconstructed tomograms are severely degraded by the physics of the measurement: the electron dose must be kept low to limit radiation damage, and the specimen can be tilted only over a restricted angular range, so part of Fourier space is never measured at all (a slice from a tomogram is shown in Figure~\ref{fig:1}A).
Macromolecules are buried in noise and surrounded by dense, unlabelled cellular material.
At the same time, the structures themselves are largely known, with over 200,000 experimentally determined entries in the RCSB Protein Data Bank~\citep{burley_updated_2025} and AlphaFold~\citep{bertoni_alphafold_2026} predicted models covering most of the proteome.
The problem we address is connecting the two: given a tomogram, identify which known structures are present and where.
Supervised learning on real tomograms is not a route to this, as annotations mostly cover large, abundant protein types and are missing precisely the small and rare proteins that are hard to find in increasingly larger datasets.
We therefore revert to simulation as the source of training data at the scale required. Because most components of the forward model of a cryo-ET experiment are well characterised, this is a viable strategy.

In this paper, we propose using the forward model to generate physically-grounded and domain-specific augmented paired views of the same scene and additional information from the simulator to learn representations invariant to nuisance variables.
For example, masked modelling (e.g. masked autoencoders~\citep{he_masked_2022} or the patch-masking objective in DINOv2~\citep{oquab_dinov2_2024}) involves tuning the mask ratio as a hyper-parameter and randomly masking patches. In contrast, the mask here is not a design choice: it is the missing wedge, a specific anisotropic region of Fourier space imposed by the microscope and determined by the acquisition parameters. As proteins have arbitrary orientations, each protein loses a different part of its spectrum due to the missing wedge.

We incorporate these ideas into the LeJEPA self-supervised learning (SSL) framework~\citep{balestriero_lejepa_2025}, yielding CARNIVAL (\textbf{C}ryo-ET \textbf{A}nnotation from \textbf{R}epresentations trained for \textbf{N}uisance-\textbf{I}nvariance via \textbf{V}iews \textbf{a}nd \textbf{L}abels), an encoder trained only on simulated tomograms that maps tomogram blocks to dense spatial features.
Using a benchmark dataset of real tomograms with ground-truth annotations, we show that CARNIVAL's embeddings enable superior detection and classification compared with TomoTwin~\citep{rice_tomotwin_2023}, a state-of-the-art model also trained on simulated data but without forward model-based paired views or privileged information.

\section{Physically-grounded views and privileged labels}
\label{sec:methods}

\begin{figure}[t]
  \centering
  \includegraphics{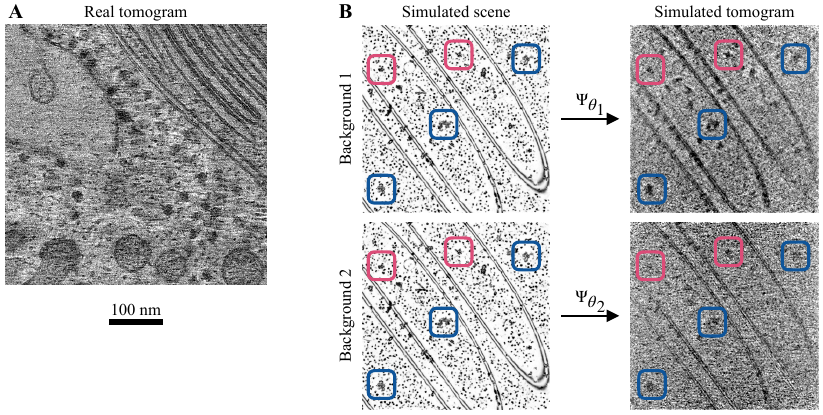}
  \caption{
    \textbf{A} Slice of real tomogram from \citep{kelley_toward_2026}.
    \textbf{B} From one arrangement of proteins and membranes we generate six views (four shown). Two clean views (left) share those structures and differ only in the small-protein background. Each corrupted view (right) is obtained by applying the measurement operator $\Psi$ to the clean view in the same row, at its own parameter draw $\theta_1$ or $\theta_2$ (e.g. tilt range, contrast transfer function, coloured noise).
    Blue and pink boxes mark two protein types; the same colour marks the same type in all four panels.
    Scale bar 100\,nm is common to all panels.
  }
  \label{fig:1}
\end{figure}

To leverage a possibly imperfect simulator of the image formation process, we generate
three sources of information that real data alone cannot provide, used jointly during
training:

\begin{enumerate}
  \item \textbf{Paired clean and corrupted views of the same specimen: the physics axis.} Corruption due to the measurement process, applied in Fourier space (e.g. missing wedge). Real data has no clean counterpart. This defines the invariance the model learns (rows in Figure~\ref{fig:1}B).

  \item \textbf{Paired views sharing structure positions but not orientations or background: the content axis.} The same proteins are embedded in a different background soup and with different orientations, with diversity due to the specimen rather than the instrument: a second view under a nuisance not covered by the measurement model (left column in Figure~\ref{fig:1}B).

  \item \textbf{Complete scene labels.} Each protein's position and identity is known, so positives are built between distinct blocks that share identity. This is privileged information in the sense of \citep{vapnik_new_2009} (available during training, absent at deployment), and is what lets the objective operate on features read out at protein locations (protein labels in Figure~\ref{fig:1}B).
\end{enumerate}

Together, these three sources make an imperfect simulator an asset rather than a liability. Two principles govern how we use it.
\emph{Invariance, not imitation}: the model is never asked to reproduce a simulated density, only to recognise what corrupted versions of one have in common.
For example, the biology is the part of our simulator most obviously wrong (fixed library of structures, arbitrary crowding statistics, no real cellular context), which constrains how we use it: the content axis supplies a nuisance to be discarded, not a target to be matched. Therefore, measurement corruptions supply training signal in the form of invariances to be learned (with respect to such corruptions).
Simulator error reaches the representation only insofar as it changes what survives corruption, so it is still important that the simulator captures the nuisance distribution accurately enough.
\emph{Coverage, not calibration}: the parameters governing corruption strength are not calibrated per dataset. Ranges are chosen so that plausible experimental configurations fall inside the sampled bounds, with the aim of making a single model usable across acquisition settings without per-dataset calibration dependent on a perfect simulator. 
This resembles domain randomisation~\citep{tobin_domain_2017}, except that the paired
views make randomisation over forward model parameters the invariance target, not just input diversity.

\subsection{Paired views}
\label{sec:views}

Let $s$ be a 3D volume representing the structures of interest in the biological specimen: proteins at known positions $\{p_j\}$ and orientations $\{\omega_j\}$ with identities $\{c_j\}$, membranes, and other large-scale structures. 
We model the process of obtaining a tomogram $x$ using the forward operator $\Psi_{\theta}$,
$
  x = \Psi_{\theta}(s + b),
$
where the parameter vector $\theta$ encapsulates all image acquisition, reconstruction and processing parameters, and $b$ is an independent instance of the random cellular background environment, a soup of small proteins below the detection limit and filling the intracellular space that generates a noisy and crowded background.
We then write $\Psi_{\theta} = R_{\theta_R} \circ M_{\theta_M}$ with $\theta = [\theta_R, \theta_M]^T$,
the composition of a measurement operator $M_{\theta_M}$ that represents the physics of the image acquisition process and outputs the collected tilt images, and a reconstruction operator $R_{\theta_R}$ that takes the tilt images and outputs the reconstructed tomogram, including all the processing steps applied during and after the reconstruction (e.g. filtering, denoising).

The three clean and three corrupted views $v_{\text{clean}_i}$, $v_{\text{corrupted}_i}$ used to train the model are
\begin{equation*}
  v_{\text{clean}_i} := s_i + b_i, \qquad
  v_{\text{corrupted}_i} := \Psi_{\theta_i}(v_{\text{clean}_i}), \qquad
  \text{for} \qquad i = 1, 2, 3,
\end{equation*}
where $s_1 = s_2 = s$, and $s_3$ keeps the positions $\{p_j\}$ and identities $\{c_j\}$ of $s$ but redraws the orientations $\{\omega_j\}$.
The first two views draw their own parameters and background, $\theta_i \sim p(\theta)$ and $b_i \sim q(b)$ for $i = 1, 2$, while the third reuses those of the first, 
$\theta_3 = \theta_1$ and $b_3 = b_1$, so that it differs from $v_1$ in protein orientation alone.
Examples of clean and corrupted views for $i=1,2$ are shown in Figure~\ref{fig:1}B.

The measurement operator $M_{\theta_M}$ is further decomposed into a sequence of transformations, each with their own parameters that we sample from their distributions during training.
For example, to simulate the tilt images, the maximum tilt angle $\theta_{\max}$ is sampled from $\{45, 50, 55, 60\}^{\circ}$, the tilt increment $\Delta \theta$ from $\{2, 3, 5\}^{\circ}$, the defocus is sampled from $[1, 5]\mu m$, and the noise colour exponent from $[0, 1.2]$.
The reconstruction operator $R_{\theta_R}$ is the standard weighted backprojection algorithm with a regularisation parameter applied in Fourier space, whose value is sampled during training.
The full list of transformations with their ranges is given in Appendix~\ref{apx:all transformations}, Table~\ref{tab:transformations all}.
Both the clean data generation (Appendix~\ref{apx:training data}) and the forward model transformations are applied on-the-fly (implemented using PyTorch and run on GPU), enabling the generation of clean and corrupted views.

\subsection{Identity loss across scenes}
\label{sec:training}

CARNIVAL consists of an encoder backbone $E$, a readout network $F$, and an optional segmentation head $G$, trained jointly.
$E$ maps a simulated volume $x$ to a dense spatial feature volume $y=E(x)$ at half the input resolution. 
For a protein at position $p$ in $x$, the local patch centred at $p$ in the feature volume $y$ is mapped by the readout network to a 512-dimensional vector
$z(x, p) = F\big(\mathrm{crop}_p(E(x))\big)$.
To fully leverage the multi-protein blocks with known protein positions and identity made possible by the simulator, the loss 
$\mathcal{L}_{\text{scene}} = \frac{1}{|\mathcal{P}_{\mathrm{scene}}|} \sum_{\mathcal{P}_{\mathrm{scene}}} \|z - z'\|^2$
is applied to all pairs 
$(z_i, z'_j) \in \mathcal{P}_{\mathrm{scene}}$, the set of embeddings
obtained from crops of proteins with distinct positions but the same identity.
The full loss is
$\mathcal{L} = \mathcal{L}_{\text{view}}
  + \lambda_{\mathrm{s}} \mathcal{L}_{\text{scene}}
  + \lambda_{\mathrm{r}} \mathcal{L}_{\mathrm{SIGReg}}
  + \lambda_{\mathrm{g}} \mathcal{L}_{\mathrm{seg}}$.
LeJEPA~\citep{balestriero_lejepa_2025} is $\mathcal{L}_{\text{view}}$ (same scene, different view) plus the regularisation $\mathcal{L}_{\mathrm{SIGReg}}$, which is applied on batch subsets with no repeating proteins. 

The paired views of Section~\ref{sec:methods} (physics and content axes) are consumed by $\mathcal{L}_{\mathrm{view}}$, and the scene labels by $\mathcal{L}_{\mathrm{scene}}$ and by $\mathcal{L}_{\mathrm{seg}}$, an auxiliary loss for training a segmentation head using protein and membrane masks from the clean data.

\section{Experiments}
\label{sec:experiments}

\begin{figure}
  \centering
  \includegraphics{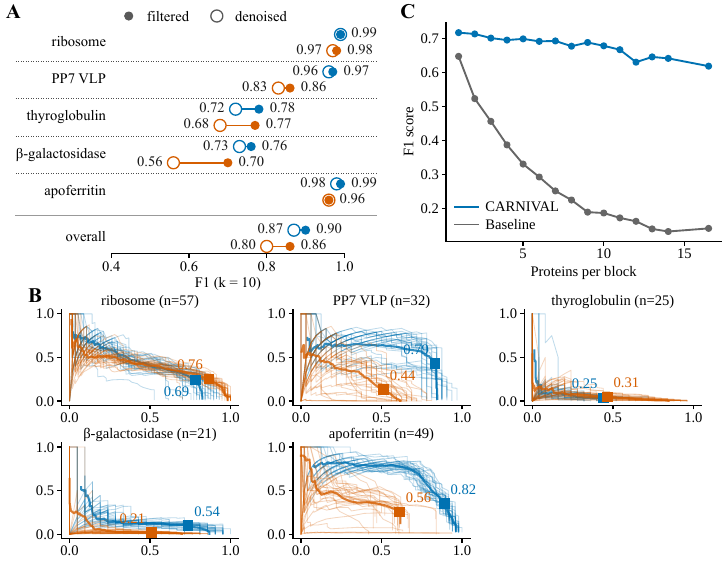}
  \caption{
    \textbf{\carnivalkey{}} vs \textbf{\tomotwinkey{}}:
    \textbf{A} Classification $F_1$ score, per protein type and overall (averaged over protein types). CARNIVAL outperforms TomoTwin on filtered tomograms (overall $0.90$ vs $0.86$) with a lower drop on denoised tomograms compared to TomoTwin ($0.87$ vs $0.80$).
    \textbf{B} Detection precision-recall on one filtered tomogram (\texttt{TS\_5\_2}) containing five protein types. Faint curves are individual template draws (5 templates/draw, 30 draws), bold curves their mean. The square marks the best $F_4$ over the threshold sweep per method.
    \textbf{C} As the number of proteins in a simulated block increases, CARNIVAL's $F_1$ stays high while the baseline degrades. 
  }
  \label{fig:2}
\end{figure}

We evaluate CARNIVAL on a real dataset with no finetuning. We use the public test set from the CZII CryoET Object Identification challenge~\citep{peck_lessons_2026}, which contains 121 real tomograms with a mixture of six annotated protein types. The organisers of the challenge note the low confidence in the quality of the annotations for one protein type, so we only include the other five types, with a total of 15,226 protein instances.
The portal provides each tomogram in two processing forms, which we use as deposited: a band-pass filtered reconstruction and one that has additionally been denoised.
We compare CARNIVAL with TomoTwin~\citep{rice_tomotwin_2023}, also trained on simulated data but without paired views.
For the \textbf{classification} and \textbf{crowding} experiments, which evaluate the quality of the embeddings at their ground-truth coordinates, we report the standard $F_1$ score (harmonic mean of precision and recall). For the \textbf{detection} experiment, we follow the Object Identification challenge~\citep{peck_lessons_2026} and report the $F_4$ score, which weights recall 16 times over precision (preferable in the case of rare proteins).

\paragraph{Classification} To evaluate the quality of the learned embeddings, we test how well they can discriminate between different protein types, given the proteins' ground-truth positions in tomograms.
We perform leave-one-out nearest neighbour classification with $k=10$ neighbours using the proteins from all tomograms and report the $F_1$ score for each model.
The results are shown in Figure~\ref{fig:2}A: CARNIVAL outperforms TomoTwin on all protein types, with an overall $F_1$ score (averaged over protein types) of $0.90$ on the filtered tomograms vs TomoTwin's $0.86$. On denoised tomograms, CARNIVAL's performance drops less than TomoTwin's, to $0.87$ compared to $0.80$.
TomoTwin's workflow prescribes unprocessed tomograms (i.e. not denoised). The denoised condition is outside its intended use, but we include it to show CARNIVAL's robustness on the available tomograms in the portal. The model performs well out-of-the-box on diverse processing conditions with no finetuning.

\paragraph{Detection} We evaluate detection on three tomograms of varying protein composition, each in its filtered and denoised version. 
We follow TomoTwin's reference-based protocol, replacing its two manual choices --- reference selection and distance threshold --- with five references randomly drawn per type over 30 draws and a sweep over the threshold to compute precision-recall curves, so the results do not depend on dataset-specific tuning.
For each type, we report the best $F_4$ score over the sweep (an upper bound available to neither method at deployment) and summarise a tomogram by the weighted mean over its evaluated types (weights given in Appendix~\ref{apx:detection more}).
Table~\ref{tab:full f4} reports the result: CARNIVAL leads in all six conditions.
Figure~\ref{fig:2}B shows the per-type precision-recall curves for the filtered tomogram \texttt{TS\_5\_2}: while TomoTwin leads on ribosome and thyroglobulin, CARNIVAL's margins where it leads are much larger
(VLP: $0.79 \pm 0.03$ vs $0.44 \pm 0.18$, $\beta$-galactosidase: $0.54 \pm 0.08$ vs $0.21 \pm 0.08$, apoferritin: $0.82 \pm 0.03$ vs $0.56 \pm 0.15$;
$\pm$ is one standard deviation across the 30 template draws), 
and on the denoised version of the same tomogram CARNIVAL leads on 4 out of 5 types, with the fifth type's difference within the standard deviation (Appendix~\ref{apx:detection more}, Figure~\ref{fig:4}).
Per-type curves and the max $F_4$ table are given in Appendix~\ref{apx:detection more}.

\begin{table}
  \caption{Detection results: weighted mean of the best $F_4$ over the threshold sweep for each evaluated protein type in that tomogram (five, two and two respectively).
    Per-type values, with their variability across template draws, are given in Table~\ref{tab:per type f4}.
  }
  \label{tab:full f4}
  \centering
  \begin{tabular}{ccccc}
    \toprule
    Tomogram & \multicolumn{2}{c}{Filtered} & \multicolumn{2}{c}{Denoised} \\
    \cmidrule(lr){2-3}\cmidrule(lr){4-5}
     & \textbf{\carnivalkey{}} & \textbf{\tomotwinkey{}} & \textbf{\carnivalkey{}} & \textbf{\tomotwinkey{}} \\
    \midrule
    \texttt{TS\_5\_2}  & \textbf{0.553} & 0.397 & \textbf{0.537} & 0.374 \\
    \texttt{TS\_14\_1} & \textbf{0.705} & 0.527 & \textbf{0.609} & 0.518 \\
    \texttt{TS\_43\_9} & \textbf{0.805} & 0.676 & \textbf{0.700} & 0.682 \\
    \bottomrule
  \end{tabular}
\end{table}

\paragraph{Crowding} While the paired views and the $\mathcal{L}_{\text{view}}$ loss force invariance to corruptions, we argue that robustness to crowding is achieved by training using blocks with multiple proteins and $\mathcal{L}_{\text{scene}}$ applied to embeddings from crops in the feature volume at positions which correspond to proteins with the same identity -- this is made possible by labels from the simulator during training.
To test this claim, we compare an earlier version of CARNIVAL with a baseline model that has the same encoder $E$ but is trained using a more standard contrastive approach: one protein per block, using paired simulated views but without the $\mathcal{L}_{\text{scene}}$ loss, and the readout network $F$ is applied to the entire feature block $E(x)$ instead of a crop.
 Rather than isolating a single variable, this experiment compares two coherent designs.
 We evaluate the discriminative power of the two models' embeddings in crowded scenes using a simulated dataset of volumes with increasing number of proteins, and report the result in Figure~\ref{fig:2}C.
CARNIVAL's $F_1$ score stays high, while the baseline model degrades rapidly with density, supporting the idea that $\mathcal{L}_{\text{scene}}$ applies localisation pressure to the spatial features, improving the embedding quality in high-density tomograms.

\section{Discussion}
\label{sec:discussion}

The proposed strategy has two fundamental limitations, neither specific to cryo-ET.
Firstly, coverage protects against mis-set ranges, not missing corruption axes.
Widening a sampled range guards against a parameter whose real value is not estimated accurately, but if real data varies along a dimension the forward model does not capture, no width helps and nothing signals the omission.
Secondly, the corruption axes are not equally trustworthy.
The missing wedge follows exactly from acquisition geometry and the CTF is a well-established physical model, but radiation damage and noise colour are empirical approximations, whose particular forms we chose partly due to their simplicity and runtime, so that views could be generated on-the-fly on GPU during training.
An approximate corruption model risks over-invariance, where the model discards variation the measurement still carries.
Our results do not indicate how to best sample an unreliable corruption axis.

Finally, while we showed that CARNIVAL outperforms a state-of-the-art model on a real dataset, we need to fully evaluate its capabilities across a variety of experimental and processing conditions, using more comprehensive benchmarks such as POPSICLE~\citep{schwartz_popsicle_2026}. Moreover, the question of sampling an unreliable corruption axis will need to be explored further by quantifying the contribution of each axis in the simulation pipeline with targeted ablations.

\begin{ack}

We thank Jake Grimmett, Toby Darling and Ivan Clayson for help with high-performance computing. 
This work was supported by the Medical Research Council, as part of United Kingdom Research and Innovation (UKRI) [MC\_UP\_A025-1013 to S.H.W.S. and MC\_UP\_1201/31 to T.A.M.B.] and the European Commission under a Marie Skłodowska-Curie Postdoctoral Fellowship [101212334 to B.T.].
The authors acknowledge the use of resources provided by the Isambard-AI National AI Research Resource (AIRR). Isambard-AI is operated by the University of Bristol and is funded by the UK Government’s Department for Science, Innovation and Technology (DSIT) via UK Research and Innovation; and the Science and Technology Facilities Council [ST/AIRR/I-A-I/1023].

\end{ack}


{

\bibliography{paper.bbl}

}


\appendix

\section{Related work}
\label{apx:related work}

\paragraph{Self-supervised learning and augmentations}
Self-supervised vision models learn representations by requiring two views of the same image to map to nearby points in the latent space, and differ mainly in how collapse is avoided, for example through negatives (SimCLR~\citep{chen_simple_2020}), teacher-student asymmetry (DINOv2~\citep{oquab_dinov2_2024}), or an explicit distributional prior on the embeddings~(LeJEPA~\citep{balestriero_lejepa_2025}).
The augmentations they use include random crops, flips, colour jitter and blur, chosen because they preserve the semantic content of natural images.
Moreover, masked modelling (masked autoencoders~\citep{he_masked_2022} and the patch-level objective of DINOv2~\citep{oquab_dinov2_2024}) involves hiding part of the input and learning to predict it, with the mask ratio (e.g. 75\,\% for masked autoencoders) tuned as a hyper-parameter and the masked patches drawn at random.
In our approach, the augmentations are given by the measurement itself (tomogram corruptions), and the mask is the missing wedge, applied in Fourier space rather than to input patches, determined by the tilt angles used during acquisition (here $\pm 45^{\circ}$ to $\pm 60^{\circ}$).
As proteins have arbitrary orientations, each loses a different part of its spectrum due to the same wedge.

\paragraph{Latent representations of physical systems} 

Joint-embedding objectives are increasingly applied to physical systems: as surrogates for expensive simulation~\citep{he_learning_2019}, as representations of spatiotemporal fields~\citep{qu_representation_2026}, and as world models whose latents are shown to encode physical structure~\citep{maes_leworldmodel_2026}.
There the paired views are related by time evolution and the latent is judged by whether it supports prediction. 
A cryo-ET specimen is frozen, so the forward operator takes the place of the transition function. 
It is the known physics that generates the corrupted views, but it projects the scene through an instrument instead of advancing it in time, and the latent is judged by what identity information survives.

\paragraph{Annotation models for cryo-ET}
Many deep learning-based annotation methods in cryo-ET are fully supervised and are trained on hand-curated particle lists or segmentations, for example crYOLO~\citep{wagner_sphire-cryolo_2019}, DeepFinder~\citep{moebel_deep_2021}, DeePiCt~\citep{de_teresa-trueba_convolutional_2023} and Easymode~\citep{so-last_easymode_2026}.
CARNIVAL belongs to the category of models that learn an embedding space in which particles of the same type are neighbours, and annotate by reference rather than by retraining. TomoTwin~\citep{rice_tomotwin_2023} is the earliest of these, trained with a triplet loss on subvolumes of simulated tomograms.
Other recent methods include MiLoPYP~\citep{huang_milopyp_2024}, AffinityVAE~\citep{famili_affinity-vae_2025}, CryoSIAM~\citep{stojanovska_cryosiam_2025}, and ProPicker~\citep{wiedemann_propicker_2026}.
None of these methods use the forward model as the corruption that generates the paired views their invariance objective is applied to: where two views are used, they are related by generic image augmentations (intensity jitter, cropping, low and high-pass filtering, voxel dropout and additive noise) rather than by independent draws of the acquisition parameters.
Moreover, those trained on simulated data generate it in advance and store it, which bounds how much of the parameter space a fixed dataset can cover.
Our simulation runs on GPU inside the training loop, so new clean scenes are generated frequently (every 50 batches), and corrupted tomograms are generated using independent parameter draws for every batch, giving the pairing that the objective needs for invariance.

\paragraph{Domain randomisation}
Randomising simulator parameters so that real data is captured by the distribution of the training data was introduced for transferring robotic policies from simulation in~\citep{tobin_domain_2017}, and recently applied to brain MRI data in~\citep{billot_robust_2023}.
In the context of cryo-ET, domain randomisation with simulated data is used in~\citep{harastani_template_2025}, where a crowded scene is generated by inserting foreground proteins and background particles at random positions, tilt images are simulated with the physics-based simulator Parakeet~\citep{parkhurst_parakeet_2021} and tomograms are reconstructed with IMOD~\citep{kremer_computer_1996}, then it is shown that DeepFinder re-trained on this simulated data improves over the baseline model.
In the same spirit, we generate clean crowded scenes (Appendix~\ref{apx:training data}) which we use to simulate diverse tomograms.
We then further use domain randomisation to provide an invariance target:
because the clean scene is held fixed while the parameters are redrawn, each pair of views tells the model which variation to discard, and so additional supervision is provided by the pairs.
This idea is also employed in~\citep{dey_learning_2025}, where two views synthesised from the same randomly generated label map are used to train a nuisance-invariant general-purpose model for 3D biomedical volumes.
The views share the geometric augmentations and only differ in appearance.
Our views are instead related by the forward model of the microscope, so the corruption is sampled in the space of acquisition parameters rather than of image effects. This also gives every corrupted view a clean counterpart and adds a second, biological axis of variation (different crowded cellular environment and different protein orientations) that appearance randomisation does not provide.

\section{Model architecture and training}
\label{apx:encoder structure}

\begin{figure}
  \centering
  \includegraphics{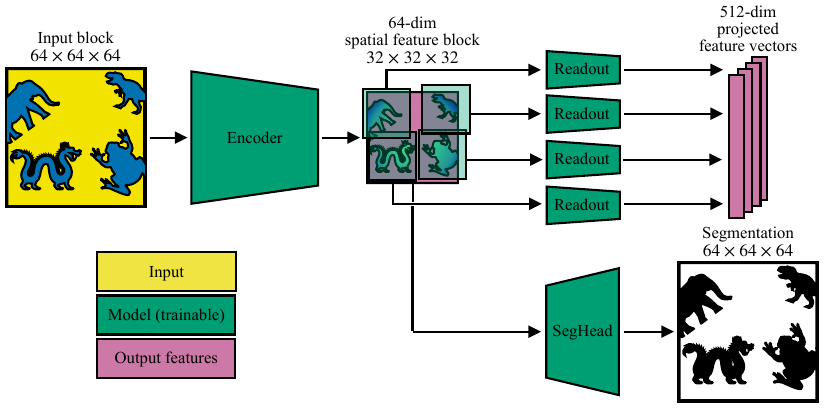}
  \caption{Model architecture.}
  \label{fig:model architecture}
\end{figure}

The encoder $E$ is based on the latent diffusion encoder~\citep{rombach_high-resolution_2022} adapted to 3D with two downsampling steps, and consists of ResNet blocks and an attention block at the final level before the output, followed by a 2x upsampling block consisting of interpolation, convolution and ResNet blocks and a skip connection from the layer before the second downsampling, resulting in spatial features that are downsampled by a factor of 2 with respect to the input volume resolution.
$E$ maps $64^3$ input volumes $x$ extracted from a tomogram to $64 \times 32^3$ dense spatial feature volumes $y=E(x)$ with $64$ channels.
The readout network $F$ consists of a convolutional block and an MLP block. $F$ is only used in training, and it maps a $64 \times 16^3$ volume, cropped from the spatial feature volume and centred at protein position, to an embedding $z \in \mathbb{R}^{512}$ that the loss function in Section~\ref{sec:training} consumes.
An optional segmentation head, consisting of an upsampling block, ResNet blocks and a skip connection from the layer before the first downsampling in the encoder, maps the spatial feature volume $y=E(x)$ to a segmentation mask $2 \times 64^3$ at the original input volume resolution, with two channels: one for proteins and one for membranes.
The segmentation loss $\mathcal{L}_{\mathrm{seg}}$ uses ground-truth masks obtained by thresholding the simulated proteins and membranes.
The full model has 9,708,706 parameters and all the components are trained jointly.
A schematic of the architecture is shown in Figure~\ref{fig:model architecture}.

The model was trained using the \texttt{AdamW} optimiser in PyTorch with learning rate of \texttt{5e-5} and \texttt{weight\_decay=5e-4}, \texttt{betas = (0.9, 0.999)} and \texttt{eps = 1e-8}, with 1000 warmup steps and cosine annealing with minimum learning rate of \texttt{5e-8}. 
The weights of the loss terms defined in Section~\ref{sec:training} were $\lambda_{\mathrm{s}} = 1.0$, $\lambda_{\mathrm{r}} = 0.1$, and $\lambda_{\mathrm{g}} = 0.1$, and $\mathcal{L}_{\mathrm{SIGReg}}$ was computed using $4096$ slices and $41$ quadrature points for the computation of the Epps-Pulley test.

The model sees six different versions of each batch sample (a $64^3$ volume): first, one clean scene containing proteins and membranes ("big structures") is sampled every 50 batches, which is used to generate two distinct clean volumes with the same big structures but different crowded background. A third clean tomogram is generated with the same protein identities and positions but in different orientations. For each batch, the first two clean tomograms are used to generate a corrupted tomogram with different parameters from Table~\ref{tab:transformations all} (see Section~\ref{sec:views} and Appendix~\ref{apx:training data}), while the third clean tomogram reuses the first view's draw, and the batch elements are extracted from the same positions from each of the resulting six tomograms (with a small jitter of up to $\pm 9$ voxels for each view): three clean views and three corrupted views for the same scene. On each GPU, a batch size of 8 was used (i.e. 8 different positions in a tomogram), each of six views (three clean and three corrupted). As the model was trained on one node with six GPUs and gradient accumulation over 2 steps, the effective batch size was $96$, with six views per batch element.

The model was trained from scratch for 100,000 steps with all transformations in Table~\ref{tab:transformations all}, with a cosine annealing learning rate schedule.
The final checkpoint was used to generate all results in this manuscript, except the crowding experiment which used an earlier version of the model.
The training of this model was done on a node with $6 \times $ NVIDIA GeForce RTX 5090 GPU with 32GB memory each, for approximately 25 days for 100,000 steps, while intermediate models during development were trained on the Isambard-AI cluster~\citep{mcintosh-smith_isambard-ai_2024}.

\section{Training data}
\label{apx:training data}

While a future goal is to train CARNIVAL using most proteins in the RCSB Protein Data Bank~\citep{burley_updated_2025} and the AlphaFold Protein Structure Database~\citep{bertoni_alphafold_2026}, the model described in this paper has been trained on the 112 PDB IDs from TomoTwin's training set larger than 100kDa~\citep{rice_tomotwin_2023}, and downloaded from the Protein Data Bank.

First, for each protein in the list above, the atomic model was downloaded from the Protein Data Bank, which was then used to compute a volume containing the protein's density by placing Gaussians at the atom coordinates in a $256^3$ box with 2\AA/voxel, low-pass filtered to 20\AA\space resolution and rescaled to a $64^3$ box with 10\AA/voxel. We refer to these as the \textit{foreground proteins}.
An additional set of 1000 $32^3$ volumes containing small proteins of up to 30kDa, whose atomic models were randomly selected from the PDB, are computed using the same steps. These are used to generate the random crowded environment of the tomograms, so we will refer to these as the \textit{background proteins}.

Every 50 batches (25 optimiser steps), foreground proteins sampled from the volumes above are inserted in a slab of random thickness between 120 and 220 voxels, within a box of dimensions $350 \times 350 \times 350$, at random positions sampled using Poisson disk sampling, together with a simulated membrane.
The membrane is obtained by tiling a simulated membrane patch along a curved surface with parameters randomly drawn.
Using this configuration of foreground structures, three clean views that correspond to the same scene are created: in the first two, the foreground proteins are in the same (random) orientations, but are surrounded by different random background protein configurations, also sampled using Poisson disk sampling, using a smaller radius and taking into account the foreground proteins and the membrane to avoid overlaps. The third clean view contains the same foreground proteins in different orientations, and the same background as the first clean view. 
For each batch, two corrupted tomograms are computed using the first two clean views after sampling the parameters for each transformation in Table~\ref{tab:transformations all}, and a third corrupted tomogram is obtained from the third clean view using the transformation parameters of the first view. This results in six total views at 10\AA/voxel: three clean and three corrupted.
Finally, \texttt{batch\_size} volumes are extracted from each of the six views from the same positions across views, with a small random jitter in each view. Each GPU performs this independently using disjoint sets of \texttt{PDB\_IDs}. We ensure that each volume contains at least one protein, and also return a list of protein coordinates and identities present in each volume and the corresponding foreground protein and membrane masks.

\section{Full list of transformations in the measurement operator}
\label{apx:all transformations}

All transformations used to generate the physically-grounded tomogram augmentations for training are given in Table~\ref{tab:transformations all}. These are used as described in Section~\ref{sec:views}, Appendix~\ref{apx:encoder structure} and Appendix~\ref{apx:training data}.

\begin{table}
  \caption{The sampled operators in the forward model $\Psi_{\theta} = R_{\theta_R} \circ M_{\theta_M}$, grouped by the two blocks of the decomposition. 
    Every simulated tomogram draws its own values (one draw per tomogram, shared by all \texttt{batch\_size} blocks extracted from it) and each range is chosen to
    bracket experimental practice rather than to match any one dataset. 
    Held fixed: $10$\,\AA/voxel, $300$\,kV, $C_s = 2.7$\,mm, amplitude contrast $0.1$, and radiation damage, which is physics rather than a processing choice and so is never switched off.
    The CTF correction-state randomises a processing choice rather than a physical one, as different reconstruction algorithms differ in their CTF treatment.
    The notation used in each row is independent from the rest of the paper.
  }
  \label{tab:transformations all}
  \centering
  \setlength{\tabcolsep}{6pt}
  \begin{tabular}{@{}p{0.18\textwidth}p{0.50\textwidth}p{0.26\textwidth}@{}}
    \toprule
    Transformation & Form & Sampled from \\
    \midrule
    \multicolumn{3}{@{}l@{}}{\emph{Measurement operator} $M_{\theta_M}$} \\
    \addlinespace
    Bulk medium \newline (ice / cytoplasm)
      & $v \mapsto v + \phi\,(v_{\max} - v_{\min})\,d(v)$ on the volume before projection, with
        $d(v) = \mathrm{clamp}(1 - v/v_{\max},\, 0,\, 1)$ so structure displaces medium; fill
        level $\phi = \alpha$
      & $\alpha \in [0.05, 0.45]$ \\
    \addlinespace
    Large-scale medium \newline variation
      & $\phi$ generalised to a smooth field $\phi(x) = \alpha\,(1 + r\,z(x))$ of the same mean,
        $z$ zero-mean unit-std low-passed by $e^{-2\pi^{2}\sigma^{2} f^{2}}$,
        $\sigma = \ell / \text{pixel size}$
      & $r \in [0, 0.5]$ \newline
        $\ell \in [400, 3200]\,\text{\AA}$ \\
    \addlinespace
    Tilt scheme \newline (missing wedge)
      & projections at $\theta \in \{-\theta_{\max}, \dots, \theta_{\max}\}$ in steps of
        $\Delta\theta$; the wedge follows from $\theta_{\max}$
      & $\theta_{\max} \in \{45, 50, 55, 60\}^{\circ}$ \newline
        $\Delta\theta \in \{2, 3, 5\}^{\circ}$ \\
    \addlinespace
    Defocus (CTF)
      & $\hat{I} \mapsto C(k; \Delta f_{\theta})\,\hat{I}$ per tilt, with $\Delta f_{\theta}$
        varying with depth along the beam
      & $\Delta f \in [1, 5]\,\mu\mathrm{m}$ \\
    \addlinespace
    CTF envelope
      & $E(k) = \exp(-B k^{2} / 4)$ from partial coherence, tilt-independent, multiplied into the
        CTF
      & $B \in [0, 300]\,\text{\AA}^{2}$ \\
    \addlinespace
    Per-tilt contrast decay
      & tilt $\theta$ scaled by $\exp\!\big(\!-C\,(\sec\theta - 1)\big)$, the Beer--Lambert
        fan-out; after CTF and before noise, so the noise floor stays tilt-independent
      & $C \in [0, 1.2]$ \\
    \addlinespace
    Radiation damage
      & $i$-th tilt image scaled by \newline $g_i(f) = \exp\!\big(\!-D_i / (2 N_c(f))\big)$ before
        noise, with $N_c(f) = 0.245 f^{-1.665} + 2.81$ and $D_i$ accumulated dose by tilt $i$ under dose-symmetric ordering
        \citep{grant_measuring_2015}
      & $D_{\max} \in [80, 120]\,\mathrm{e}/\text{\AA}^{2}$\\
    \addlinespace
    Noise level
      & Gaussian noise added to every tilt with standard deviation
        $\sigma\,(I_0^{\max} - I_0^{\min})$, referenced to the contrast range of the zero-tilt
        frame
      & $\sigma \in [0, 0.25]$ \\
    \addlinespace
    Noise colour
      & that noise weighted by $w(f) \propto (f + \varepsilon)^{-\gamma/2}$
        renormalised to $\langle w^{2} \rangle = 1$ so total variance is unchanged
      & $\gamma \in [0, 1.2]$ \\
    \midrule
    \multicolumn{3}{@{}l@{}}{\emph{Reconstruction operator} $R_{\theta_R}$} \\
    \addlinespace
    CTF correction state
      & transfer applied $\in \{1,\; C E,\; |C| E\}$: amplitude-restored, uncorrected,
        phase-flipped
      & uniform over the three \\
    \addlinespace
    Dose weighting
      & the same $g_i(f)$ re-applied inside backprojection, now to signal $+$ noise, as the
        matched filter that merges the frames; otherwise plain WBP
      & applied with probability $0.5$ \\
    \addlinespace
    Backprojection \newline regularisation
      & Fourier-space division $1 / (W(k) + \lambda)$, with $W$ the accumulated tilt weights
      & $\lambda \in \{1, 2, 5, 10\}$ \\
    \bottomrule
  \end{tabular}
\end{table}

\section{Experiments: further details}
\label{apx:experiments}

We use the public test set from the CZII CryoET Object Identification challenge~\citep{peck_lessons_2026} (downloaded from the CryoET Data Portal~\citep{ermel_data_2024} deposition ID CZCDP-10310, DS-10445), which provides each tomogram in two processing forms that we use as deposited: a band-pass filtered reconstruction, and one that has additionally been denoised with DenoisET~\citep{peck_aretomolive_2026}.
Tomograms are used at the $\sim\!10$\,\AA{} voxel spacing at which the model was trained. 
No resampling or finetuning is applied; each tomogram is only clipped and z-scored, as in training.
The same CARNIVAL checkpoint (Appendix~\ref{apx:encoder structure}) is used for the classification and detection experiments.
The crowding experiment compares an earlier pair of models and is described separately below.

All three experiments report an $F_\beta$ score:
\begin{equation*}
  F_\beta = (1 + \beta^2)\,\frac{\mathrm{Precision} \times \mathrm{Recall}}
                               {\beta^2 \times \mathrm{Precision} + \mathrm{Recall}},
\end{equation*}
computed per protein type and then averaged over types. 
Precision and Recall are defined as:
\begin{equation*}
  \text{Precision} = \frac{\text{TP}}{\text{TP}+\text{FP}}
  \qquad
  \text{Recall} = \frac{\text{TP}}{\text{TP} + \text{FN}},
\end{equation*}
where TP is the number of true positives, FP is the number of false positives and FN is the number of false negatives (i.e. missed particles).

The classification and crowding experiments evaluate embeddings at ground-truth positions, so every particle receives exactly one predicted label. In this case, precision and recall are symmetric and we use the standard $F_1$, their harmonic mean.
Detection has no such symmetry. The operating point is a distance threshold, and lowering it trades false positives (which a later classification or averaging step can still reject) against particles that are missed outright and cannot be recovered downstream. 
We therefore follow the Object Identification challenge~\citep{peck_lessons_2026} and report $F_4$, which weights recall $\beta^2 = 16$ times more heavily than precision, both for that reason and so that our numbers are comparable to the challenge's.

For the comparison with TomoTwin~\citep{rice_tomotwin_2023}, we used the TomoTwin model version \texttt{09.2023} made available by the authors on Zenodo at the URL \url{https://zenodo.org/records/8358240}, under a Creative Commons Attribution 4.0 International licence.

All the experiments presented in this article were run on an internal cluster node with $6 \times $ NVIDIA GeForce RTX 5090 GPU with 32GB memory each, AMD EPYC 9354 32-Core Processor and 512GB RAM, for a total execution time of approximately 12 hours.

\subsection{Classification}
\label{apx:classification more}

For a protein at position $p$ in the input tomogram, a $64^3$ block around $p$ is encoded, and the embedding is the concatenation of the feature vectors of $E(x)$ within a sphere of radius 4 feature voxels (8 tomogram voxels) around $p$.
The concatenated vectors of all particles in a dataset are L2-normalised and reduced to 64 dimensions by PCA, fitted without labels on those same particles, and cosine distance is the metric in this space.
For TomoTwin, the embedding is its 32-dimensional output for the $37^3$ subvolume centred on $p$, also compared by cosine distance.
This test asks only whether the embedding space separates protein types. The positions are given, so detection is assumed and what is measured is the geometry of the embeddings themselves.
Proteins of the same type should fall in the same region of the embedding space regardless of which tomogram they came from, their orientation, or how much surrounding material is in their neighbourhood.

Embeddings from all 121 tomograms of one processing type are pooled and classified by leave-one-out $k$-nearest-neighbour vote with $k=10$ and uniform weights, each particle being classified by its neighbours across the whole pooled set with itself excluded. 
Per-type $F_1$ scores and the macro average over the five types are reported in Table~\ref{tab:classification}, and are the same numbers plotted in Figure~\ref{fig:2}A.

\begin{table}
  \caption{Full classification results with $k=10$ nearest neighbours.
  The $F_1$ score is shown for each protein type and the overall score is averaged over protein types. These are the numbers shown in Figure~\ref{fig:2}A.}
  \label{tab:classification}
  \centering
  \begin{tabular}{cccccc}
    \toprule
    & \multicolumn{2}{c}{Filtered $F_1$} & \multicolumn{2}{c}{Denoised $F_1$} & \# particles\\
    \cmidrule(lr){2-3}\cmidrule(lr){4-5}
     & \textbf{\carnivalkey{}} & \textbf{\tomotwinkey{}} & \textbf{\carnivalkey{}} & \textbf{\tomotwinkey{}} \\
    \midrule
    ribosome              & \textbf{0.991} & 0.978 & \textbf{0.990} & 0.969 & 6478 \\
    VLP               & \textbf{0.970} & 0.860 & \textbf{0.959} & 0.830 & 726 \\
    thyroglobulin         & \textbf{0.779} & 0.773 & \textbf{0.722} & 0.682 & 1501 \\
    $\beta$-galactosidase & \textbf{0.757} & 0.705 & \textbf{0.726} & 0.561 & 827 \\
    apoferritin           & \textbf{0.986} & 0.962 & \textbf{0.978} & 0.962 & 5694  \\
    \midrule
    overall & \textbf{0.896} &  0.855 & \textbf{0.875} & 0.801  & 15226 \\
    \bottomrule
  \end{tabular}
\end{table}

\subsection{Detection} 
\label{apx:detection more}

\begin{table}
  \caption{Per-type detection results. Each entry is the maximum $F_4$ over the threshold sweep, taken on the precision-recall curve averaged over the 30 template draws, $\pm$ one standard deviation of the individual draws' $F_4$ at that same threshold.
    These are the operating points marked in Figures~\ref{fig:4}--\ref{fig:6}. 
    Bold marks the higher of the two values within a processing condition.
    The weighted-mean rows reproduce the corresponding entries of Table~\ref{tab:full f4}.}
  \label{tab:per type f4}
  \centering
  \setlength{\tabcolsep}{4pt}
  \begin{tabular}{lcccc}
    \toprule
    & \multicolumn{2}{c}{Filtered} & \multicolumn{2}{c}{Denoised} \\
    \cmidrule(lr){2-3}\cmidrule(lr){4-5}
    \qquad Protein type & \textbf{\carnivalkey{}} & \textbf{\tomotwinkey{}}
                 & \textbf{\carnivalkey{}} & \textbf{\tomotwinkey{}} \\
    \midrule
    \multicolumn{5}{l}{\texttt{TS\_5\_2}} \\
    \qquad ribosome                 & $0.691 \pm 0.076$ & $\mathbf{0.757 \pm 0.049}$ & $\mathbf{0.683 \pm 0.055}$ & $0.567 \pm 0.086$ \\
    \qquad VLP                      & $\mathbf{0.789 \pm 0.032}$ & $0.438 \pm 0.176$ & $\mathbf{0.768 \pm 0.037}$ & $0.395 \pm 0.140$ \\
    \qquad thyroglobulin            & $0.251 \pm 0.067 $ & $\mathbf{0.305 \pm 0.079}$ & $0.236 \pm 0.073$ & $\mathbf{0.240 \pm 0.082}$ \\
    \qquad $\beta$-galactosidase    & $\mathbf{0.537 \pm 0.079}$ & $0.205 \pm 0.080$ & $\mathbf{0.533 \pm 0.079}$ & $0.225 \pm 0.089$ \\
    \qquad apoferritin              & $\mathbf{0.817 \pm 0.025}$ & $0.565 \pm 0.145$ & $\mathbf{0.770 \pm 0.044}$ & $0.728 \pm 0.169$ \\
    \addlinespace
    \quad Weighted mean            & $\mathbf{0.553}$ & $0.397$ & $\mathbf{0.537}$ & $0.374$ \\
    \midrule
    \multicolumn{5}{l}{\texttt{TS\_14\_1}} \\
    \qquad ribosome                 & $\mathbf{0.836 \pm 0.022}$ & $0.821 \pm 0.017$ & $\mathbf{0.775 \pm 0.027}$ & $0.666 \pm 0.052$ \\
    \qquad apoferritin              & $\mathbf{0.575 \pm 0.051}$ & $0.233 \pm 0.147$ & $\mathbf{0.442 \pm 0.131}$ & $0.370 \pm 0.222$ \\
    \addlinespace
    \quad Weighted mean            & $\mathbf{0.705}$ & $0.527$ & $\mathbf{0.609}$ & $0.518$ \\
    \midrule
    \multicolumn{5}{l}{\texttt{TS\_43\_9}} \\
    \qquad ribosome                 & $0.730 \pm 0.036$ & $\mathbf{0.804 \pm 0.049}$ & $\mathbf{0.657 \pm 0.067}$ & $0.615 \pm 0.060$ \\
    \qquad apoferritin              & $\mathbf{0.880 \pm 0.021}$ & $0.548 \pm 0.096$ & $0.743 \pm 0.066$ & $\mathbf{0.749 \pm 0.101}$ \\
    \addlinespace
    \quad Weighted mean            & $\mathbf{0.805}$ & $0.676$ & $\mathbf{0.700}$ & $0.682$ \\
    \bottomrule
  \end{tabular}
\end{table}

A feature volume is computed for each tomogram by applying the encoder $E$ in overlapping $64^3$ blocks with stride 8, keeping the central $48^3$ of each block and blending the overlaps with a smoothly decaying (Hann) window, which avoids the seams a hard crop leaves behind.
TomoTwin~\citep{rice_tomotwin_2023} was run with stride 2, as recommended in TomoTwin's documentation at \url{https://tomotwin-cryoet.readthedocs.io/en/stable/tutorials/tutorials_overview.html#embed-your-tomogram}.
Both methods produce feature volumes at half the tomogram sampling, so the candidate grids coincide and the two are scored on identical positions. 
For CARNIVAL, the embeddings at each position are obtained from this feature volume using a similar PCA procedure to the classification experiment, while for TomoTwin, the protein embeddings are vectors in the feature volume at those positions.

We evaluate detection using a workflow similar to TomoTwin's reference-based workflow~\citep{rice_tomotwin_2023}, as follows.
For each of 30 independent draws, five ground-truth particles per type are sampled as templates.
Every voxel of the feature volume is assigned to the type of its nearest template across all types, by cosine distance, and a type's detection candidates are the local minima of this distance among its assigned voxels.
Any ground-truth particle within 8 feature voxels of a template is removed from the test set for that draw, so neither the templates themselves nor their immediate neighbours can be scored as detections.
The distance at each candidate is the threshold variable: sweeping it over 100 values spanning the $0.001$st to the $10$th percentile of the distance volume traces one precision-recall curve per type per draw.
A candidate counts as a true positive when it falls within 4 feature voxels (8 tomogram voxels) of a ground-truth particle of that type; matching is one-to-one and greedy, taking candidates in order of increasing distance to their nearest unmatched ground-truth particle, so the best-localised detections are matched first.
Protein types with fewer than 5 instances left after template exclusion are dropped from the figures and from the weighted mean calculation in Table~\ref{tab:full f4}.
A type below that bar still contributes its templates to the nearest-type assignment, so a scarce type cannot inflate another type's precision by going unrepresented.

For each type we report the best $F_4$ over the sweep (an upper bound available to neither method at deployment, but computed identically for both) and summarise a tomogram by the mean over its evaluated types, weighting the hard ones ($\beta$-galactosidase and thyroglobulin) by 2 and the rest by 1.
The weights are those used to score the Object Identification challenge~\citep{peck_lessons_2026}, where the additional $\beta$-amylase type is not evaluated: the organisers did not score it because of low confidence in the quality of its annotations.

\begin{figure}
  \centering
  \includegraphics{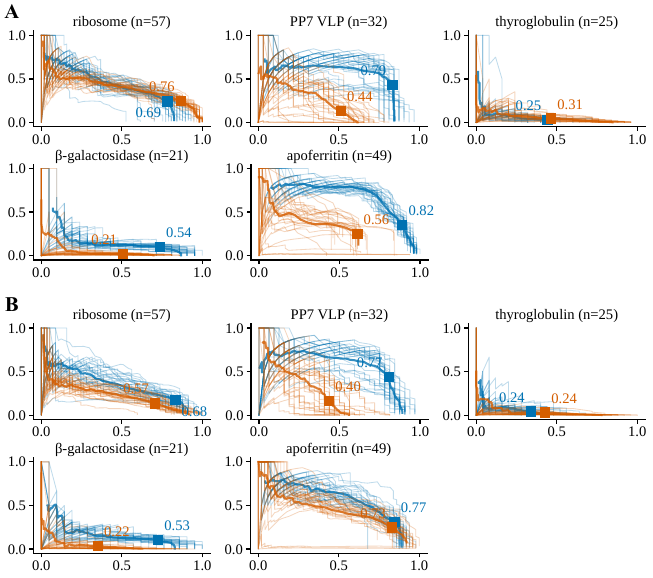}
  \caption{
    \textbf{\carnivalkey{}} vs \textbf{\tomotwinkey{}}: Detection precision-recall curves, tomogram \texttt{TS\_5\_2}. $n$ is the number of particles evaluated, after discarding the templates and their neighbours.
    \textbf{A} Filtered. This panel is identical to Figure~\ref{fig:2}B, reproduced here for easy visual comparison between the filtered and denoised tomograms.
    \textbf{B} Denoised.
  }
  \label{fig:4}
\end{figure}

\begin{figure}
  \centering
  \includegraphics{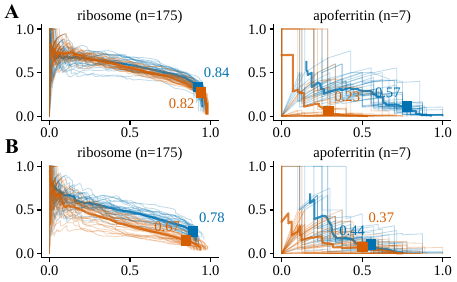}
  \caption{
    \textbf{\carnivalkey{}} vs \textbf{\tomotwinkey{}}: Detection precision-recall curves, tomogram \texttt{TS\_14\_1}. $n$ is the number of particles evaluated, after discarding the templates and their neighbours.
    \textbf{A} Filtered.
    \textbf{B} Denoised.
  }
  \label{fig:5}
\end{figure}

\begin{figure}
  \centering
  \includegraphics{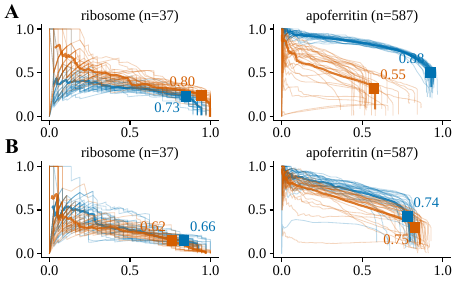}
  \caption{
    \textbf{\carnivalkey{}} vs \textbf{\tomotwinkey{}}: Detection precision-recall curves, tomogram \texttt{TS\_43\_9}. $n$ is the number of particles evaluated, after discarding the templates and their neighbours.
    \textbf{A} Filtered.
    \textbf{B} Denoised.
  }
  \label{fig:6}
\end{figure}

Per-type precision-recall curves are shown in Figures~\ref{fig:4}--\ref{fig:6}.
Figure~\ref{fig:4}A repeats the tomogram of Figure~\ref{fig:2}B beside its denoised counterpart (panel B); Figure~\ref{fig:5} and Figure~\ref{fig:6} cover two tomograms dominated by one type, ribosome or apoferritin respectively.
Table~\ref{tab:per type f4} gives the maximum $F_4$ for each protein type, taken on the draw-averaged precision-recall curve (the values marked in Figures~\ref{fig:4}--\ref{fig:6}), each with the standard deviation across the 30 template draws at that threshold.

\begin{figure}
  \centering
  \includegraphics{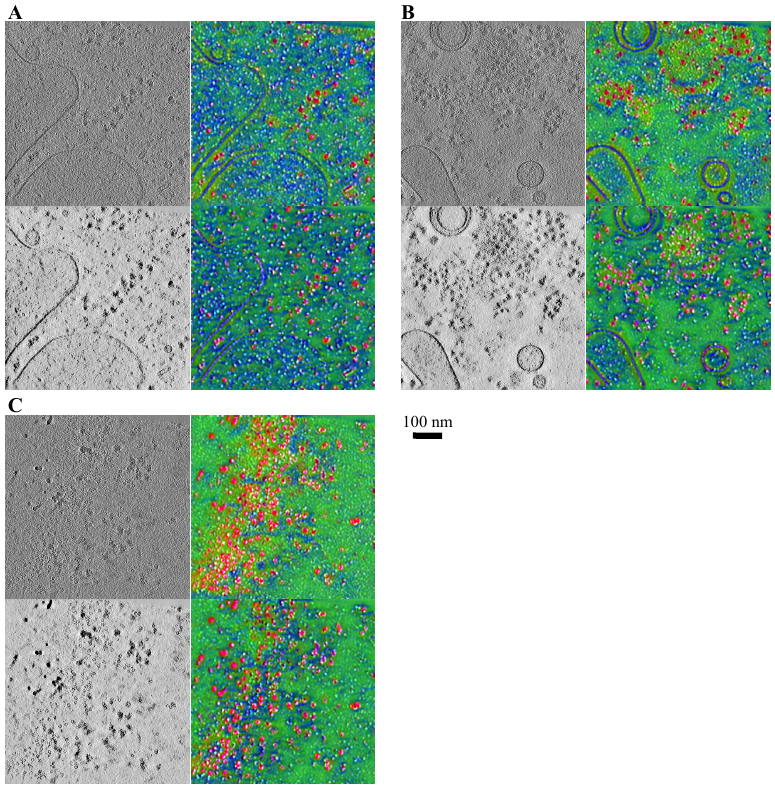}
  \caption{Slices of tomograms and spatial feature volumes from CARNIVAL, with the filtered version on the top row and the denoised version on the bottom row in each panel. As the spatial feature blocks have 64 channels, we show the 3 largest principal components of each feature vector, mapped to RGB, where the principal components were fitted once on the six displayed slices pooled together. Scale bar 100\,nm is common to all panels.
    \textbf{A} Tomogram \texttt{TS\_5\_2}.
    \textbf{B} Tomogram \texttt{TS\_14\_1}.
    \textbf{C} Tomogram \texttt{TS\_43\_9}.
  }
  \label{fig:detection tomos}
\end{figure}

The tomograms are deposited in the CryoET Data Portal under IDs TM-19018 and TM-19019 (denoised and filtered \texttt{TS\_14\_1}), TM-19098 and TM-19099
(\texttt{TS\_43\_9}), and TM-19158 and TM-19159 (\texttt{TS\_5\_2}), all in dataset DS-10445, the same public test set used for classification.
Slices of the three tomograms are shown in Figure~\ref{fig:detection tomos}.

\subsection{Crowding} 
\label{apx:crowding experiment}

This experiment is run on simulated data, where the number of proteins in a block is known exactly. 
Tomograms of $250 \times 350 \times 350$ voxels ($z \times y \times x$) are simulated with 2000 proteins drawn from the same 112-protein library used for training (Appendix~\ref{apx:training data}), at a fixed $\pm 60^{\circ}$ tilt range. 
Crowding is controlled through the minimum centre-to-centre separation of the Poisson-disk packing, swept over $\{18, 20, 22, 25, 30, 35, 40\}$ voxels.
A $64^3$ block is then extracted around each protein and its density recorded as one plus the number of other proteins falling inside it. 
A block is discarded if it shares a protein with a block already kept, so no two blocks overlap. 
Because discarding overlapping blocks results in fewer blocks from tomograms with high density, we simulated more tomograms in this regime.

The embeddings from the two models are obtained using an identical readout:
the average of the spatial feature vectors over a spherical region of radius 2 voxels at the feature volume centre.
This comparison isolates what differs in training (i.e. one vs many proteins in each input block) rather than in evaluation.
Embeddings are pooled across all densities and classified by leave-one-out $k$-nearest neighbours with $k=10$ and uniform weights, and the macro-average $F_1$ is then stratified by density.
To make the per-density comparison meaningful in Figure~\ref{fig:2}C, only proteins appearing at at least 14 distinct density levels were kept, so that every level is scored on the same set of types, and the sparsely populated levels 15-18 were merged into one.
Within a level, the macro average runs over the types actually present there.

Both models were trained for 100,000 steps using a subset of the transformations in Table~\ref{tab:transformations all} and share the same encoder backbone.
They differ as described in Section~\ref{sec:experiments}:
the baseline sees one protein per block, applies its readout to the whole feature block rather than a crop at the protein position, and is trained without $\mathcal{L}_{\text{scene}}$.
Figure~\ref{fig:2}C reports the result: CARNIVAL's $F_1$ stays high across the density range while the baseline degrades with density, which is what the localisation pressure of $\mathcal{L}_{\text{scene}}$ is meant to prevent.

\clearpage
\newpage
\input{checklist.tex}

\end{document}

%% file: checklist.tex
\section*{NeurIPS Paper Checklist}

\begin{enumerate}

\item {\bf Claims}
    \item[] Question: Do the main claims made in the abstract and introduction accurately reflect the paper's contributions and scope?
    \item[] Answer: \answerYes{}
    \item[] Justification:
        We show in Section~\ref{sec:experiments} that our model, CARNIVAL, outperforms a state-of-the-art model, TomoTwin, on two different tasks: classification of protein types at their ground-truth positions, and detection of proteins in tomograms using TomoTwin's template-based workflow. 

\item {\bf Limitations}
    \item[] Question: Does the paper discuss the limitations of the work performed by the authors?
    \item[] Answer: \answerYes{}
    \item[] Justification:
        We discuss in Section~\ref{sec:discussion}, Discussion, the fundamental limitations of our approach: a forward model must capture the main corruption types present in the image acquisition process it simulates; and it is not clear how one should sample the parameters of a corruption axis that is not faithfully modelled without risking over-invariance. 
        Moreover, we highlight specific limitations of our presentation in this manuscript, namely that we should evaluate our model on more comprehensive benchmarks in the future, and that ablations isolating individual components should be performed to understand exactly which contribute most to the results.

\item {\bf Theory assumptions and proofs}
    \item[] Question: For each theoretical result, does the paper provide the full set of assumptions and a complete (and correct) proof?
    \item[] Answer: \answerNA{}
    \item[] Justification:
        The paper does not include theoretical results.

    \item {\bf Experimental result reproducibility}
    \item[] Question: Does the paper fully disclose all the information needed to reproduce the main experimental results of the paper to the extent that it affects the main claims and/or conclusions of the paper (regardless of whether the code and data are provided or not)?
    \item[] Answer: \answerYes{}
    \item[] Justification: 
        In addition to the main ideas behind CARNIVAL and its training objective described in Section~\ref{sec:methods} in the main text, in the Appendix we give detailed descriptions of the model and training (Appendix~\ref{apx:encoder structure}), clean tomogram generation using simulated protein densities (Appendix~\ref{apx:training data}), the full transformations we apply to the clean tomograms to obtain realistic corrupted tomograms (Appendix~\ref{apx:all transformations}), and the complete experimental setup for our main results (Appendix~\ref{apx:experiments}).

\item {\bf Open access to data and code}
    \item[] Question: Does the paper provide open access to the data and code, with sufficient instructions to faithfully reproduce the main experimental results, as described in supplemental material?
    \item[] Answer: \answerNo{}
    \item[] Justification:
        While we do not provide code and trained models with our submission to this workshop, we are in the process of submitting an extended version of this manuscript to an archival venue. With this extended manuscript, we will publish all the code, training scripts, model weights, and scripts for reproducing all experiments, including those described in the current manuscript.

\item {\bf Experimental setting/details}
    \item[] Question: Does the paper specify all the training and test details (e.g., data splits, hyperparameters, how they were chosen, type of optimizer) necessary to understand the results?
    \item[] Answer: \answerYes{}
    \item[] Justification: 
        We describe the experimental setup in the main text in Section~\ref{sec:experiments}, where we give enough detail for the reader to understand and appreciate the results shown in Figure~\ref{fig:2} and Table~\ref{tab:full f4}.
        Full details about the experimental setup are given in Appendix~\ref{apx:experiments}.


\item {\bf Experiment statistical significance}
    \item[] Question: Does the paper report error bars suitably and correctly defined or other appropriate information about the statistical significance of the experiments?
    \item[] Answer: \answerYes{}
    \item[] Justification:
        Error bars are reported for the detection experiment (Figures~\ref{fig:2}B and~\ref{fig:4}--\ref{fig:6}, Table~\ref{tab:per type f4}), the experiment whose protocol contains a random draw.
        Detection depends on which proteins are drawn as templates, so every measurement is repeated over 30 independent draws of five templates per protein type.
        Figures~\ref{fig:2}B and~\ref{fig:4}--\ref{fig:6} show each draw's precision-recall curve individually alongside their mean, so the full spread is displayed rather than summarised.
        Per-type $F_4$ values in Table~\ref{tab:per type f4} are given as $x \pm y$, where $x$ is the maximum of $F_4$ along the draw-averaged precision-recall curve and $y$ is the population standard deviation of the 30 individual draws' $F_4$ at that same threshold: a one-sigma spread, not a standard error of the mean.
        No distributional assumption is made and no confidence interval or significance test is claimed, and no reported interval extends outside $[0,1]$.
        The aggregate weighted means in Table~\ref{tab:full f4} are taken from the same draw-averaged curves; we report the per-type spreads separately rather than propagating them into the aggregate, which would require an independence assumption across protein types.
        These error bars capture variability from template selection only, for the three chosen tomograms.
        Classification (Figure~\ref{fig:2}A) is deterministic given the trained encoder: leave-one-out $k$-NN over all 15,226 annotated particles involves no sampling.
        The crowding experiment (Figure~\ref{fig:2}C) is evaluated on a single simulated dataset (Appendix~\ref{apx:crowding experiment}) and was not repeated over independent simulations.
        Both models are scored on exactly the same blocks, so the comparison between them is paired, and each density level aggregates many blocks.


\item {\bf Experiments compute resources}
    \item[] Question: For each experiment, does the paper provide sufficient information on the computer resources (type of compute workers, memory, time of execution) needed to reproduce the experiments?
    \item[] Answer: \answerYes{}
    \item[] Justification:
        The compute resources used to run the experiments are described in Appendix~\ref{apx:experiments}, and those used to train the model in Appendix~\ref{apx:encoder structure}.

\item {\bf Code of ethics}
    \item[] Question: Does the research conducted in the paper conform, in every respect, with the NeurIPS Code of Ethics \url{https://neurips.cc/public/EthicsGuidelines}?
    \item[] Answer: \answerYes{}
    \item[] Justification:
        The research presented in this article has been conducted fully in compliance with the NeurIPS Code of Ethics.

\item {\bf Broader impacts}
    \item[] Question: Does the paper discuss both potential positive societal impacts and negative societal impacts of the work performed?
    \item[] Answer: \answerNA{}
    \item[] Justification:
       The work is a method for annotating cryo-electron tomography data, and its foreseeable consequences are confined to the practice of structural biology. 
       We identified no direct positive or negative societal impacts beyond this specialised setting, and so the paper does not include a broader impacts discussion.

\item {\bf Safeguards}
    \item[] Question: Does the paper describe safeguards that have been put in place for responsible release of data or models that have a high risk for misuse (e.g., pre-trained language models, image generators, or scraped datasets)?
    \item[] Answer: \answerNA{}
    \item[] Justification:
        The output of this research does not pose any risk of misuse.

\item {\bf Licenses for existing assets}
    \item[] Question: Are the creators or original owners of assets (e.g., code, data, models), used in the paper, properly credited and are the license and terms of use explicitly mentioned and properly respected?
    \item[] Answer: \answerYes{}
    \item[] Justification:
        We compare our model with TomoTwin, which we cite throughout the paper (both main text and appendix).
        In Appendix~\ref{apx:experiments} we state the exact TomoTwin model version, with a URL and the name of the license.
        We also use data from the Protein Data Bank and the CryoET Data Portal, which we cite both in the main text in Section~\ref{sec:experiments} and in Appendix~\ref{apx:training data} and Appendix~\ref{apx:experiments}.

\item {\bf New assets}
    \item[] Question: Are new assets introduced in the paper well documented and is the documentation provided alongside the assets?
    \item[] Answer: \answerNA{}
    \item[] Justification:
        We do not release new assets with this paper (See Question 5 above on open access to data and code).

\item {\bf Crowdsourcing and research with human subjects}
    \item[] Question: For crowdsourcing experiments and research with human subjects, does the paper include the full text of instructions given to participants and screenshots, if applicable, as well as details about compensation (if any)? 
    \item[] Answer: \answerNA{}
    \item[] Justification:
        This paper does not involve crowdsourcing nor research with human subjects.

\item {\bf Institutional review board (IRB) approvals or equivalent for research with human subjects}
    \item[] Question: Does the paper describe potential risks incurred by study participants, whether such risks were disclosed to the subjects, and whether Institutional Review Board (IRB) approvals (or an equivalent approval/review based on the requirements of your country or institution) were obtained?
    \item[] Answer: \answerNA{}
    \item[] Justification:
        This paper does not involve research with human subjects. 

\item {\bf Declaration of LLM usage}
    \item[] Question: Does the paper describe the usage of LLMs if it is an important, original, or non-standard component of the core methods in this research? Note that if the LLM is used only for writing, editing, or formatting purposes and does \emph{not} impact the core methodology, scientific rigor, or originality of the research, declaration is not required.
    \item[] Answer: \answerNA{}
    \item[] Justification:
        LLMs have been used during this project and in the preparation of this paper for text editing and code assistance in standard ways, which does not need to be documented.

\end{enumerate}